\documentclass{article}

\usepackage{arxiv}

\usepackage[utf8]{inputenc} 
\usepackage[T1]{fontenc}    
\usepackage{hyperref}       
\usepackage{url}            
\usepackage{booktabs}       
\usepackage{amsfonts}       
\usepackage{nicefrac}       
\usepackage{microtype}      
\usepackage{graphicx}
\usepackage{natbib}
\usepackage{doi}
\usepackage{subfig}
\usepackage[english]{babel}
\usepackage{amsmath}

\title{Online Out-of-Domain Detection for Automated Driving}

\author{ {\includegraphics[scale=0.06]{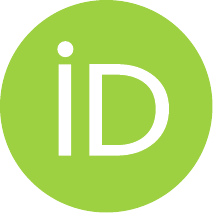}\hspace{1mm}Timo Sämann} \\
	Valeo Schalter und Sensoren GmbH\\
	Kronach Germany\\
	\texttt{timo.saemann@valeo.com} \\
	\And
	{\includegraphics[scale=0.06]{orcid.pdf}\hspace{1mm}Horst Michael Gross} \\
	Ilmenau University of Technology\\
	\texttt{horst-michael.gross@tu-ilmenau.de} \\
}

\renewcommand{\headeright}{Technical Report}
\renewcommand{\undertitle}{Technical Report}

\hypersetup{
pdftitle={Online Out-of-Domain Detection for Automated Driving},
pdfauthor={Timo Sämann, Horst Michael Gross},
pdfkeywords={Safety Requirements, DNN Insufficiency, Lack of Generalization, Autonomous Driving},
}

\begin{document}

\maketitle

\begin{abstract}
Ensuring safety in automated driving is a major challenge for the automotive industry. Special attention is paid to artificial intelligence, in particular to Deep Neural Networks (DNNs), which is considered a key technology in the realization of highly automated driving. DNNs learn from training data, which means that they only achieve good accuracy within the underlying data distribution of the training data. When leaving the training domain, a distributional shift is caused, which can lead to a drastic reduction of accuracy. In this work, we present a proof of concept for a safety mechanism that can detect the leaving of the domain online, i.e. at runtime. In our experiments with the Synthia data set we can show that a 100 \% correct detection of whether the input data is inside or outside the domain is achieved. The ability to detect when the vehicle leaves the domain can be an important requirement for certification.

\end{abstract}


\section{Introduction}
An important prerequisite for the certification of DNNs is compliance with safety requirements. The proof that the DNN meets the necessary safety requirements is required. This can be done in the form of an assurance case~\cite{burton2019confidence}.
An important part of the assurance case is the adequate mitigation of the DNN insufficiency's~\cite{samann2020strategy}. Therefore it is advisable to consider the complete machine learning life-cycle and to apply safety mechanisms that contribute to the mitigation of the DNN insufficiency. One such safety mechanism is the out-of-domain (OOD) detection at runtime. Please note that the terms out-of-domain and out-of-distribution can be used interchangeably in this work.

For a DNN, in-domain is defined by the underlying data distribution of the training data. If the test data are drawn from the same distribution, the DNN is likely to perform fairly well in terms of accuracy. For data out-of-domain, the DNN accuracy is usually significantly reduced. The DNN is not able to generalize beyond its training data distribution. Besides the increase of the generalization capability, the recognition of the domain at run-time is an important safety mechanism. The goal is to be able to recognize at runtime whether the input data are similar in their distribution to the training data and thus can be detected as \glqq in-domain\grqq{} or whether they differ strongly and thus have to be classified as \glqq out-of-domain\grqq{}. A correct detection of in- and out-of-domain can serve as a measure of uncertainty or confidence for the DNN output. If the input data is outside the domain, the output of the DNN can no longer be classified as reliable. In this case, possible measures are the use of redundancy branches that have a different in-domain due to different training data or sensor modalities or the transition to an emergency mode that terminates the application as soon as possible.

\mbox{}
\vfill
\begin{center}
Accepted at Machine Learning in Certified Systems (MLCS) Workshop, 14.-15.01.2021.\\
\end{center}

\section{Approach}
\label{sec:Approach}

We used DeepLabV3+~\cite{chen2018encoder} as baseline for semantic segmentation and extended the architecture after training with a second decoder (see green layers in Fig.~\ref{fig:pipeline}), which reconstructs the input image. When training the second decoder, all learnable parameters of DeepLabV3+ are freezed, so that the accuracy of the semantic segmentation is not affected in any way.  The approach is similar to the one in~\cite{wang2020dual} with the difference that the focus is on improving semantic segmentation. In contrast to the approach in~\cite{lohdefink2020self} the focus is on online out-of-domain detection and no additional auto-encoder is used.

The reconstruction was learned by means of the Mean Square Error (MSE) and the Kullback Leibler Divergence (KLD) loss. The training takes place in a self-supervised fashion. The KLD loss is usually measured between encoder and decoder. Since the learnable parameters of the encoder are freezed and therefore no effect of the KLD loss would exist, another convolutional layer is placed in front of the second decoder (see yellow layer in Fig.~\ref{fig:pipeline}). Both losses are weighted with a factor $\alpha$ and $\beta$ and added together. 
\begin{equation*}
\label{loss}
\textmd{Loss} = \textmd{KLD} \cdot \alpha + \textmd{MSE} \cdot \beta
\end{equation*}
In our experiments we set alpha to 0.1 and beta to 1.
With an input image size of 3x768x1280 px, the feature map size after the encoder is 256x48x80. Dividing both values results in a compression factor of 0.33.

The assumption is that the second decoder only learns certain features, so that an image similar to the training data distribution can be reconstructed. For input images that are different, the reconstruction is more difficult and leads to a higher reconstruction error. The Peak Signal-to-Noise Ratio (PSNR) is a measure of how good the reconstruction is. The PSNR is defined as follows:
\begin{equation*}
\label{loss}
\textmd{PSNR} = 10 \cdot \log\frac{1}{\textmd{MSE}}
\textmd{ with }
\textmd{MSE} = (\textmd{Input Image} - \textmd{Reconstruction\_image})^2
\end{equation*}

\begin{figure}
    \includegraphics[width=0.75\textwidth]{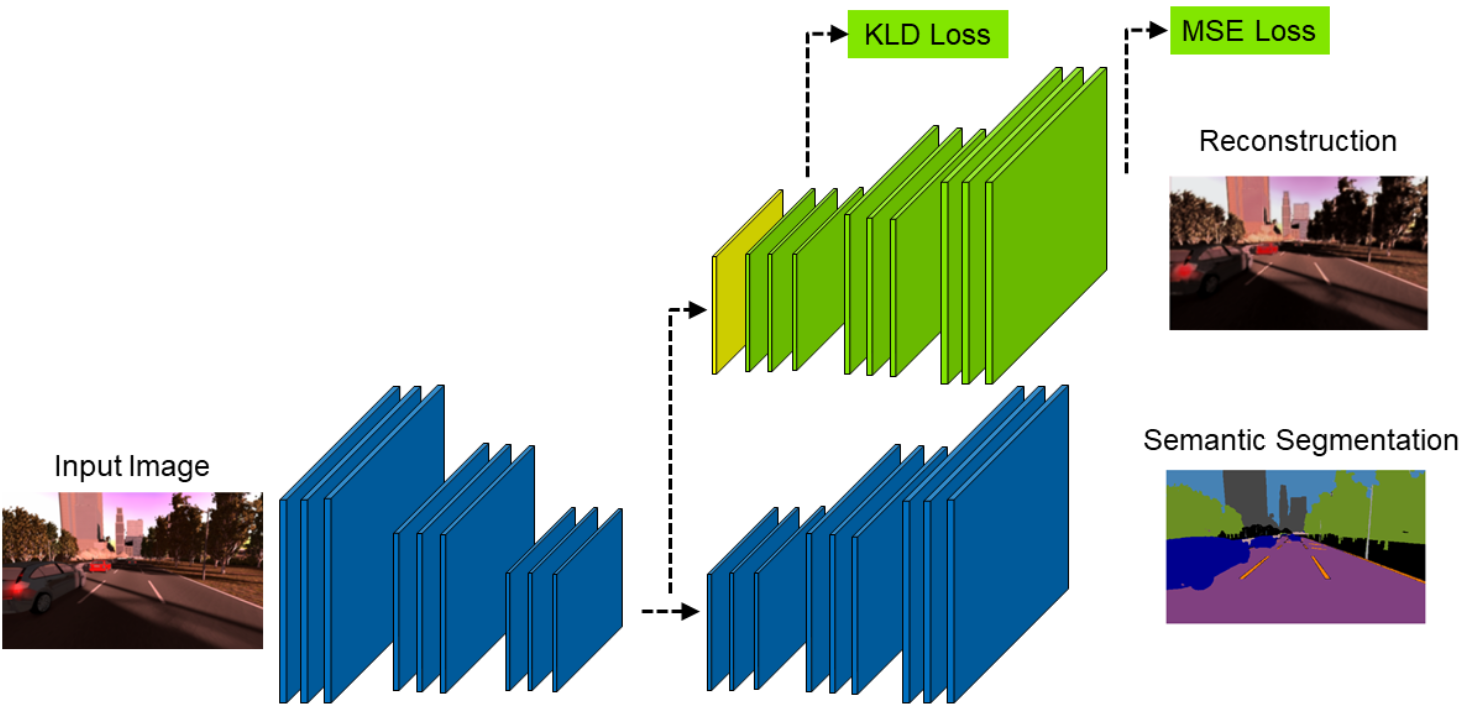}
	\centering
	\label{fig:pipeline}
\end{figure}
\begin{figure}
    \includegraphics[width=0.75\textwidth]{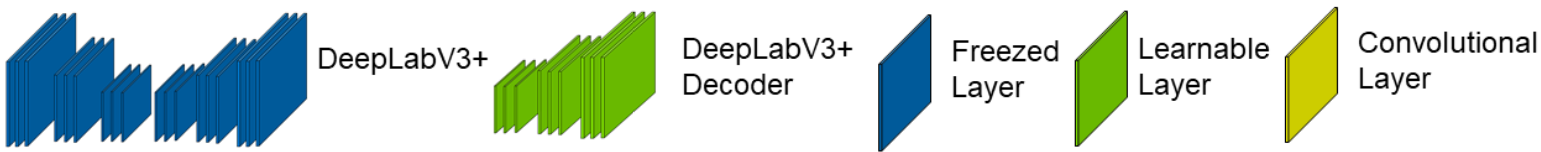}
	\centering
	\caption{Visual representation of our approach.}
	\label{fig:pipeline}
\end{figure}

\section{Dataset}
\label{sec:Dataset}

To carry out the experiments we used the Synthia Video Sequences~\cite{Ros_2016_CVPR}. It contains a large number of domains such as dawn, fog, rain, winter, summer, spring, fall, night etc. We have used 4 sequences (SEQ), 2 SEQ with Highway and 2 SEQ with New York ish. Each sequence is further divided into the domains mentioned above. 
We used sequence 1 and 2 for training and 5 and 6 for testing. 
The resolution of the Synthia data is 768, 1280 pixels. For a simple overview of the data split used, we list it in the following: 

Training data, 2392 images: Dawn (SEQ1+2).

Test data (in-domain), 1775 images: Dawn (SEQ5+6).

Test data (out-of-domain), 13921 images: Winter (SEQ5+6), sunset (SEQ5+6), summer (SEQ5+6), spring (SEQ5+6), night (SEQ5+6), fog (SEQ5+6), rain (SEQ5), rainnight (SEQ5), softrain (SEQ5), winternight (SEQ5+6).\footnote{Please note that the domains rain, rainnight and softrain are not available in sequence 6.}

\section{Results}
\label{sec:Results}
\begin{figure}%
    \centering
    \subfloat[\centering $\tau=1$]{{\includegraphics[width=7.5cm]{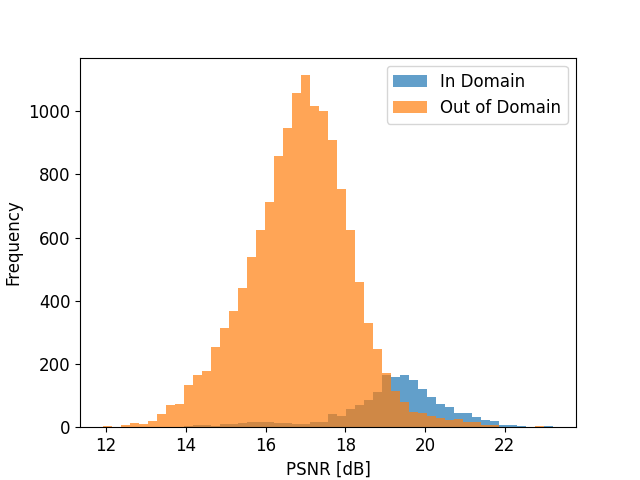} }}%
    \qquad
    \subfloat[\centering $\tau=50$]{{\includegraphics[width=7.5cm]{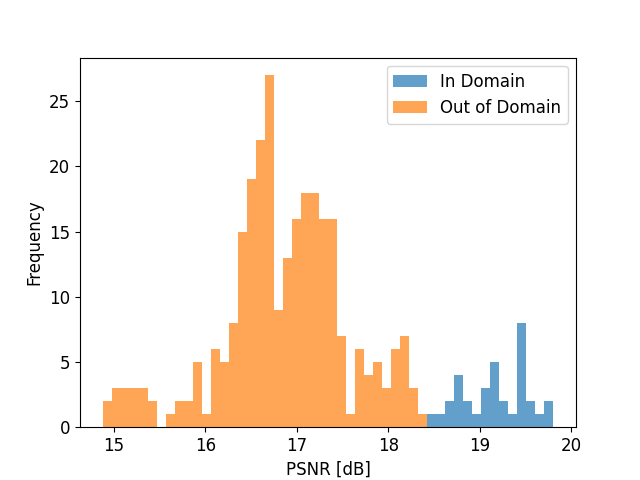} }}%
    
    \caption{Histogram for in- and out-of-domain.}%
    
    \label{fig:histo}%
\end{figure}

The PSNR for in-domain and out-of-domain data was measured. The division of the data into in-domain and out-of-domain can be read in the previous section~\ref{sec:Dataset}. The result is shown in the form of a histogram in Fig.~\ref{fig:histo}. The x-axis represents the PSNR and is divided into 50 bins. The y-axis represents the frequency, i.e. the number of images evaluated. For (a) the evaluation shows that the in-domain data are mainly between 18 and 22 dB distributed. The out-of-domain data are mainly between 13 and 20 dB. Between 18 and 20 dB a clear overlap between in-domain and out-of-domain data can be seen. Thus, a clear separation between in and out-of-domain data is not easily possible in a single image analysis. For this reason a new value $\tau$ has been introduced, which indicates how many images are combined into a sequence before a single PSNR value is determined, i.e. the average PSNR over the sequence with the length $\tau$ is determined. $\tau$ is 1 for the first histogram (a) and 50 for the second (b). This averaging of the values narrows the variance of the individual domains, since outliers at the edge of the spectrum are smoothed out. The higher $\tau$ the lower the variance. A complete separation of in- and out-of-domain data is already possible with $\tau=50$. Since the data was generated with a frequency of 5~Hz, this means that at run time every 10 seconds a reliable estimation can be made whether the input images are in or out-of-domain. At a higher frame rate, a reliable estimation can probably be made after a shorter time. 


\section{Correlation Between Reconstruction Error and DNN Accuracy}
\label{sec:Correlation}
This section describes the investigations on the correlation between the reconstruction error (PSNR) and the DNN accuracy (mean Intersection over Union, mIoU). It is assumed that for a low reconstruction error (high PSNR) a high DNN accuracy (high mIoU) is achieved and vice versa.

For the datasets in- and out-of-domain (for the used dataset split see section~\ref{sec:Dataset}) the PSNR and the mIoU were calculated per image and are shown in Fig.~\ref{sec:Correlation}. In both subplots the x-axis represents the PSNR in dB and the y-axis represents the mIoU. Each image is represented by a star. The upper subplot with the blue stars describes the in-domain data and the lower one with the yellow stars the out-of-domain data. In (a) it is clearly visible that the mIoU for out-of-domain data is drastically lower than for in-domain data. This observation was to be expected because the DNN was trained only with in-domain data.

However, a direct correlation between PSNR and mIoU without prior separation into in-domain and out-of-domain cannot be established. This means that the mIoU value cannot be deduced from the measurement of the PSNR alone. Considering only the individual subplots, it is noticeable that a minimal increase in the mIoU can be observed with increasing PSNR for in-domain. The slope of the regression line is slightly positive at 0.001. For the out-of-domain data (see subplot below) the opposite is true, i.e. the mIoU decreases with increasing PSNR.

For (b) in Fig.~\ref{sec:Correlation} the parameter $\tau=50$ is used as described in section~\ref{sec:Results}. It can be observed that with increasing $\tau$ the slope of the regression line increases. For the in-domain data this confirms the statement that with higher PSNR (i.e. a smaller reconstruction error) the mIoU (the DNN accuracy) increases.

\begin{figure}%
    \centering
    \subfloat[\centering $\tau=1$]{{\includegraphics[width=7.5cm]{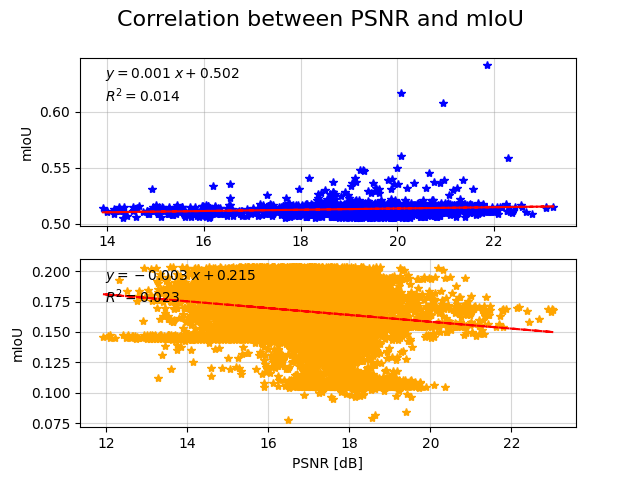} }}%
    \qquad
    \subfloat[\centering $\tau=50$]{{\includegraphics[width=7.5cm]{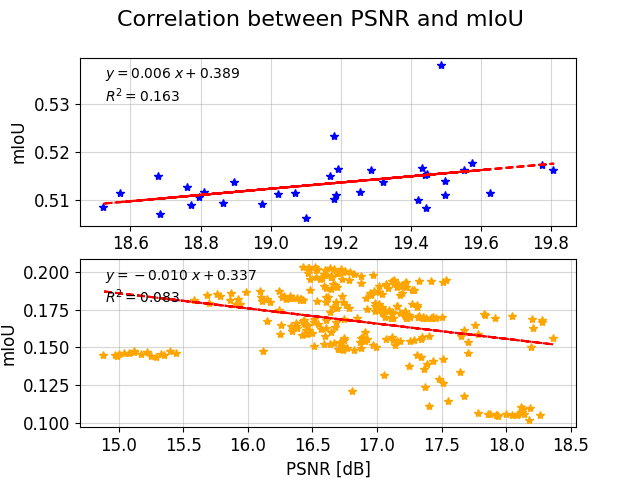} }}%
    \caption{Correlation between reconstruction error and DNN accuracy}%
    

    \label{fig:example}%
\end{figure}

\section{Conclusion}
\label{sec:Conclusion}
We have introduced a safety mechanism that can perform an out-of-domain detection at runtime. Using the Synthia data set we could show that the variance of PSNR distribution for in- and out-of-domain can be reduced by sequence evaluation. From a sequence evaluation of 50 images ($\tau=50$) a 100 \% separation between in- and out-of-domain data was achieved. The OOD detection is an important function for online monitoring of the DNN and can be a requirement for certification of DNNs.

\section*{Acknowledgement}
The research leading to these results is funded by the German Federal Ministry for Economic Affairs and Energy within the project “KI Absicherung – Safe AI for Automated Driving". The authors would like to thank the consortium for the successful cooperation. Special thanks go to Laureen Lake, Andreas Bär, Lennart Ries, Hanno Stage and Maram Akila for the fruitful discussions.

\bibliographystyle{unsrtnat}
\bibliography{references}  






\end{document}